# AgentRec: Agent Recommendation Using Sentence Embeddings Aligned to Human Feedback

**Joshua Park**
Rutgers University
josh.park@rutgers.edu

**Yongfeng Zhang**
Rutgers University
yongfeng.zhang@rutgers.edu

## Abstract

Multi-agent systems must decide which agent is the most appropriate for a given task. We propose a novel architecture for recommending which LLM agent out of many should perform a task given a natural language prompt by extending the Sentence-BERT (SBERT) encoder model. On test data, we are able to achieve a top-1 accuracy of 92.2% with each classification taking less than 300 milliseconds. In contrast to traditional classification methods, our architecture is computationally cheap, adaptive to new classes, interpretable, and controllable with arbitrary metrics through reinforcement learning. By encoding natural language prompts into sentence embeddings, our model captures the semantic content relevant to recommending an agent. The distance between sentence embeddings that belong to the same agent is then minimized through fine-tuning and aligned to human values through reinforcement learning from human feedback. This allows the classification of natural language prompts based on their nearest neighbors by measuring the cosine similarity between embeddings. This work is made possible through the generation of a synthetic dataset for agent recommendation, which we have open-sourced to the public along with the code for AgentRec recommendation system at https://github.com/joshprk/agentrec.

## 1 Introduction

While large language models are able to output convincing and coherent natural language [2, 5], they are by design unable to work beyond their predefined modalities without augmentation [8, 18, 21]. As a result, there has been an extensive effort to extend large language models by providing frameworks such as retrieval-augmented generation (RAG), chain-of-thought (COT) reasoning, and the capability to use tools [13, 24, 25]. These initiatives provide an opportunity to resolve the limitations that stand in the way of artificial general intelligence (AGI) through the development of autonomous LLM agents [1, 8, 15, 26] with domain-specific deep understanding that is expected of intelligence [14]. However, domain-specific agents only exhibit narrow intelligence with domain expertise for a specific task [4].

Therefore, a next reasonable step is to develop systems where multiple LLM agents can collaborate to solve a given task. This is the premise of the research on multiagent systems, where there has been significant research on systems where different LLM agents are given expert roles in solving a complex problem requiring reasoning abilities [12, 15]. Each LLM agent can often be described as a workflow with its own context, being granted capabilities such as chain-of-thought reasoning or the ability to use tools by outputting a specific language which triggers the tool [8]. However, most research that has been done on this subject often execute these agents in a rigid order and can only allow for a certain class of questions as there is no robust way of flexibly determining which agents out of a large selection of unrelated agents are most suited to perform the given task [1, 12, 26].

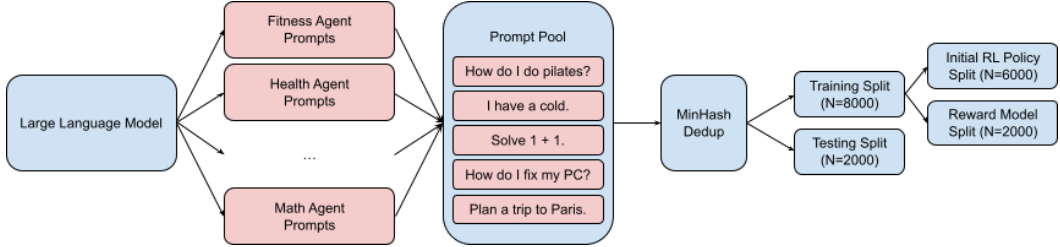

Figure 1: A summary of the synthetic data generation pipeline described in section 2.

To solve this issue, we contribute AgentRec—a scalable, fast, and efficient method of selecting an agent to perform a task given a natural language prompt. By using sentence embeddings, we are able to extract the semantic content of a natural language prompt relevant to selecting an agent and compare it to a labeled corpus of synthetically generated prompts that specific agents are expected to be able to answer in under 300 milliseconds for 8 agents with a top-1 test accuracy of 92.2%.

## 2 The Dataset

For both training and recommendation, AgentRec requires a large representative dataset of natural language sentence prompts for each agent. Due to the expensive nature of real-world data, we opted to instead use a synthetic dataset which we generated from `Llama-3.1-8B-Instruct` [9] on a NVIDIA RTX A5000. Our main challenges in creating a synthetic dataset representative of prompts primarily centered around non-repetitiveness [10]. Language models with less parameters, such as `Llama3.1-3B-Instruct` and `Llama3.1-1B-Instruct`, struggled to even generate coherent natural language. To reduce the complexity of the prompts and therefore the difficulty of synthetic generation, only single sentences were generated, which our architecture takes into consideration.

The first technique we used to reduce repetition is top-$k$ sampling, where we set the probability of any candidate out of the top $k = 50$ likely next tokens in the LLM to 0 and re-scaled the top candidates to a probability distribution [7]

$$\tau_k(p)_i = \frac{(i \in K)p_i}{\sum_{j \in K} p_j}$$

Afterwards, we apply nucleus sampling to the remaining token distribution, where only a set of the most likely tokens that sum up to a $p = 0.95$ probability are kept [11] such that

$$\sum_{x \in V^{(p)}} P(x \mid x_{:i-1}) \geq p$$

Finally, a repetition penalty of $r = 1.2$ and a temperature of $T = 0.6$ was introduced. Once the prompts were generated, the dataset was de-duplicated using a MinHash deduplication pipeline [3] to reduce potential over-fitting. The dataset represented 8 agents of varying general topics with $1,250$ prompts each, making the total number of prompts $10,000$. The following agents are: (1) tech support agent, (2) cooking agent, (3) math agent, (4) gaming agent, (5) therapy agent, (6) reading agent, (7) health agent, and (8) fitness agent. Some agents were selected to be as unrelated as possible from other agents—such as the cooking agent—whereas some agents were selected to have a soft overlap of their expertise domain such as the health and fitness agents.

These datasets were randomly shuffled and split uniformly into a training split ($N = 8000$) and a testing split ($N = 2000$) such that every agent had an equal number of prompts in each split. The training split was then further split such that the base encoder finetuning dataset ($N = 6000$) and the reward model dataset ($N = 2000$) do not overlap. Finetuning the base model allows us to produce the initial policy model for which to perform reinforcement learning from human feedback (RLHF) [16]. The base encoder finetuning dataset was re-organized before training such that it generated a (anchor, positive, negative) triplet from the single sentence dataset through the `sentence_transformers` Python library's `BatchAllTripletLoss` function. We chose to rigorously split the dataset to ensure that semantic content relevant to recommending an agent is truly learned and that any part of our architecture does not overfit to the training dataset.

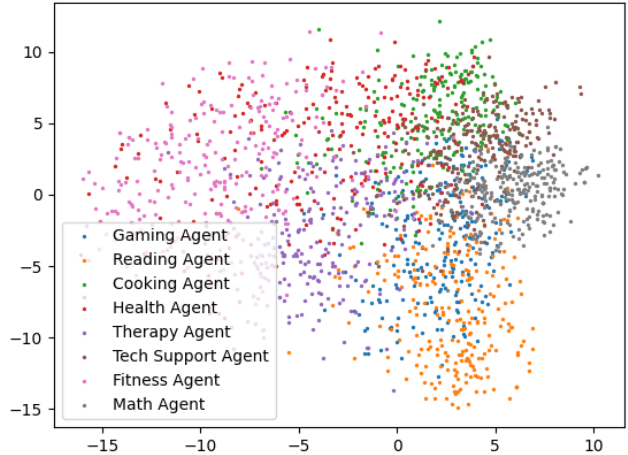

Figure 2: Sentence embeddings from the test split ($N = 2000$) generated from the `all-mpnet-base-v2` foundational model reduced from 768 dimensions to 2 dimensions through principal componenent analysis (PCA).

## 2.1 Data Exploration

An initial study of this dataset using a foundational Sentence-BERT (SBERT) [19] encoder model—`all-mpnet-base-v2` [22]—to generate sentence embeddings, indicates that assigning a specific agent from a pool of agents to a given task is non-trivial to models that are not specifically trained for this problem. As shown in figure 2, using an encoder model that is not specifically trained for agent recommendation will provide embeddings that are often noisy and prone to fail at edge cases. In cases where an agent recommender does not recommend an agent which is designed to complete the task, it is likely that a chain-of-thought system or other failsafe must be used to redirect it to the appropriate agent, which can be expensive both in terms of time and computational resources [24].

Another issue is that these embeddings are not aligned to human expectations, which leaves the possibility that this model recommends in a fashion that is unexpected to human values due to the nature of the dataset format even if finetuned traditionally. Finally, it is important to note the structure of the text in the dataset. While a LLM user prompt can be of variable size and structure, the SBERT architecture was designed to work with single sentences [19]. This results in a limitation where user prompts must follow the rigid structure of specifying enough information in the very first sentence for a naive architecture to produce a meaningful recommendation.

## 3 The Architecture

Our proposed architecture extends the original Sentence-BERT model to provide an end-to-end method for recommending an agent given a natural language user prompt. The machine learning objective is to produce sentence embeddings where embeddings for a given agent generate a clean, separable cluster. To achieve this, it is possible to model the problem of agent recommendation as a sentence similarity problem. If it is possible to numerically determine which sentence corpus in the dataset described in section 2 is most similar to a given natural language user prompt, then one can recommend the agent which that given corpus represents.

Therefore, agent recommendation first starts with the generation of embedding corpora through the SBERT encoder, one for each agent. These corpora contain the embeddings of a representative set of prompts that the agent should answer. These embeddings can be cached given that the weights of the SBERT encoder being utilized do not change, allowing for fast initialization of the agent recommendation system [19]. In our tests, each of the 8 agents used 1,000 prompts from the train split.

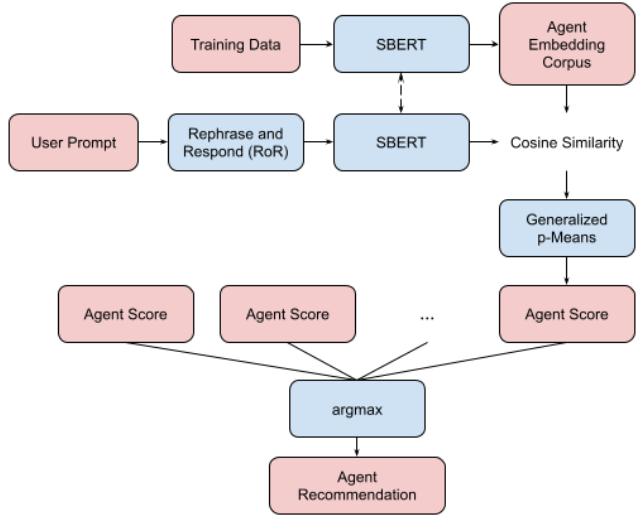

Figure 3: Summary of the AgentRec architecture.

When a user prompts the multiagent LLM system, the prompt is standardized to the sentence structure seen in the dataset through rephrase and respond (RaR) [6]. This allows the multiagent LLM system to handle variable length prompts where the first sentence does not sufficiently describe the task. For example, a prompt may first contain a set of data and have the sentence describing the task at the very end. This also provides the multiagent LLM system the opportunity to autonomously correct any spelling mistakes and to re-frame the task to better understand the human user's intent [20]. The rephrased prompt is then encoded into a sentence embedding with the same SBERT encoder used by the agent sentence embedding corpora. To find the top-$k$ best agents suited to the task, the cosine similarity of all embeddings in all corpora and the user prompt is calculated.

$$\cos(\theta) = \frac{\mathbf{A} \cdot \mathbf{B}}{\|\mathbf{A}\|\|\mathbf{B}\|} = \frac{\sum_{i=1}^{n} A_i B_i}{\sqrt{\sum_{i=1}^{n} A_i^2} \sqrt{\sum_{i=1}^{n} B_i^2}}$$

Afterwards, a score function is used which produces a final likelihood value from the cosine similarity values that can be used to find the top-$k$ agent recommendations. We found that logarithmic generalized p-means works best, which produces a score $S_q$ for a given agent $q$ such that

$$S_q = \ln \left( \frac{1}{n} \sum_{i=1}^{n} \left[ \left( \frac{\sum_{i=1}^{n} A_i B_i}{\sqrt{\sum_{i=1}^{n} A_i^2} \sqrt{\sum_{i=1}^{n} B_i^2}} \right)^p \right] \right)^{1/p}$$

$$= \frac{1}{p} \cdot \ln \left( \frac{1}{n} \sum_{i=1}^{n} \left[ \left( \frac{\sum_{i=1}^{n} A_i B_i}{\sqrt{\sum_{i=1}^{n} A_i^2} \sqrt{\sum_{i=1}^{n} B_i^2}} \right)^p \right] \right)$$

Once calculated, the $k$ highest generalized p-means represent the top-$k$ recommendations of the agent recommendation system.

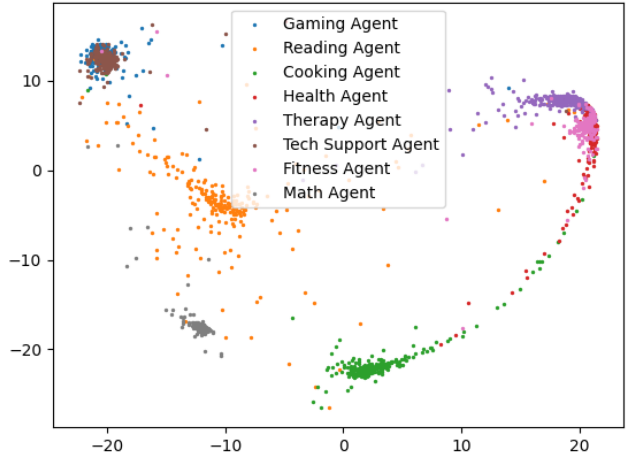

Figure 4: Sentence embeddings from the test split ($N = 2000$) generated by the AgentRec architecture reduced from 768 dimensions to 2 dimensions through principal componenent analysis (PCA).

## 3.1 Score Function

The score function used can drastically affect the accuracy rate of the agent recommendation system. For example, simply using the highest cosine similarity from an agent embedding corpus as the score for the respective agent results in a top-1 test accuracy of $32.55\%$. During preliminary testing, it was found that comparing the mean of the cosine similarity values of each corpus results in the highest top-1 test accuracies. Therefore, much of our study on score functions is focused on measurements of central tendency.

Before choosing a score function, it is important to note that the range of cosine similarity values is $-1 \leq x \leq 1$. Therefore, any exponentiation of numbers farther from the extremes of the range will reduce the impact of moderate numbers closer to 0, whereas multiplication will increase the impact of moderate numbers.

A naive score function using the arithmetic mean of each agent embedding corpus produces reasonable results with a top-1 test accuracy of $90.05\%$. However, using geometric mean—which weighs smaller numbers more heavily—produces a top-1 test accuracy of $61.05\%$. The pattern becomes much clearer when using generalized p-means with a high $p$-value as the score function. In generalized p-means, a larger $p$-value increases the effect of larger numbers in the mean. Accurate agent recommendation relies heavily on cosine similarity scores which the recommendation system can confidently assess as similar (+1) or dissimilar (-1). With a value of $p = 200$—which accentuates the extreme cosine similarity scores to a high degree—we were able to produce a top-1 accuracy rating of $92.2\%$.

## 3.2 Reinforcement Learning from Human Feedback

As shown in section 2.1, a foundational SBERT encoder model which is not trained for the explicit purpose of agent recommendation is ill-equipped to perform the task. Furthermore, even with traditional finetuning, it is irresponsible to claim that a model trained naively on the dataset described in section 2 is aligned to human values [17]. To resolve this issue, we utilized RLHF to produce an initial RL policy from supervised finetuning (SFT) and aligned an initial policy to human values through a reward model. On a NVIDIA RTX A5000, the entire training pipeline takes around 20 minutes to complete with the dataset described in section 2.

## 3.3 Evaluation Challenges

One interesting finding was that a higher learning rate in supervised finetuning without RLHF increased the top-1 test accuracy of the overall system, but had glaringly obvious edge cases. Using a learning rate of $1 \times 10^{-4}$, a finetuned SBERT encoder had a top-1 test accuracy of 93.55%. However, it was unable to accurately recommend cases such as "how do I eat well?" appropriately to the health agent, instead recommending the fitness agent to answer the prompt. A limitation of a lack of real-world data and statistics on how often certain prompts are actually asked to multiagent LLM systems prevent us from assessing the real-world impact of our methodology. One difficult-to-classify edge case may be asked much more frequently than many simpler prompts, and our data deduplication pipeline would prevent that from being reflected in the test accuracy.

Table 1: Selected prompts showing an ability to differentiate nuances in tasks for similar domains

| Prompt | Recommended Agent |
| --- | --- |
| How do I see a doctor? | Health Agent |
| How do I eat well? | Health Agent |
| How do I sleep well? | Therapy Agent |
| How do I deal with calculus in my teeth? | Health Agent |
| How do I go for a run? | Fitness Agent |

## 4 Results

AgentRec provides a robust framework for agent recommendation, with a top-1 test accuracy of 92.2% and resilience against edge cases which require an understanding of the nuances between prompts that are structurally similar but semantically different. This property is shown by the recommendations in table 1, where prompts that all relate to the general theme of health and wellness are assigned to agents which should have deeper understanding of the task's targeted subtopic. For example, the question of sleeping well is assigned to the therapy agent, which deals more with mental wellbeing, instead of the health agent that deals with general wellbeing. In our testing, we found that recommendations generally take less than 300 milliseconds to complete per prompt on a single NVIDIA RTX A5000 given that all agent corpora embeddings are cached and all model parameters are loaded into memory. Without rephrase and respond, the time to complete drops to 50 milliseconds per prompt on average after CPU warm-up, although it is likely that this number can be decreased even further with an efficient generalized p-means calculation using logarithmic estimations to prevent numerical instability.

Furthermore, it is robust to potential differences in the definition of the same word in different contexts due to the transformer architecture of the SBERT encoder [23]. This property is visible when the prompt "how do I deal with calculus in my teeth?" is classified by AgentRec, where it recommends the health agent instead of recommending the math agent as it did in a naive SBERT-only architecture.

A 2-dimensional PCA visualization of the test split embeddings show that our training methodology results in the agent corpora becoming clustered in the embedding space in figure 4 in comparison to the noisy and overlapping space shown in figure 2. The separable clustering of the agent corpora suggests that agent recommendation is a task that can benefit from study in comparison to the naive approach of using the original SBERT model.

## 5 Conclusion

In this work, we presented the unexplored task of agent recommendation in multiagent systems, provided an initial framework of recommending as a data-driven learning task, and implemented a recommendation system using these principles to build a highly accurate agent recommendation system.

Our method is fast, scalable, and efficient, leveraging the computational efficiency of SBERT siamese architecture through caching sentence embeddings. We are interested in the potential of multiagent LLM systems as a means to study the problem of building an artificial general intelligence that has deep understanding of many different domains of expertise.

The code we used for the synthetic dataset generation, as well as training and testing the AgentRec architecture is available at https://github.com/joshprk/agentrec.

## A   Additional Visualizations of Test Embeddings

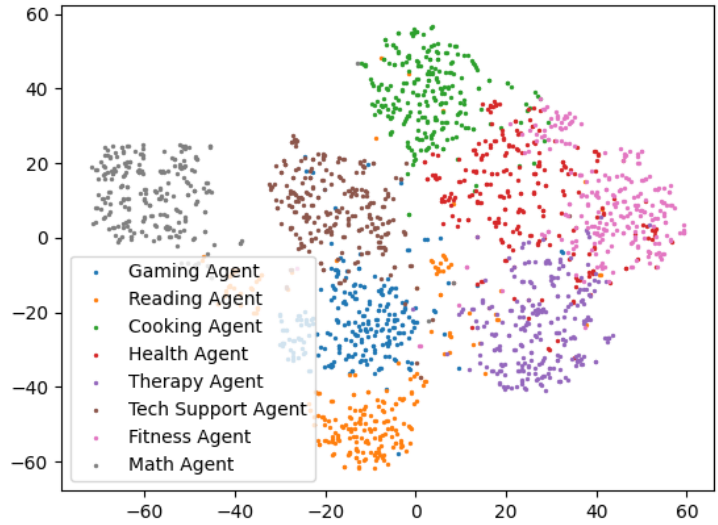

Figure 5: Sentence embeddings from the test split ($N = 2000$) generated by `all-mpnet-base-v2` reduced from 768 dimensions to 2 dimensions through t-distributed shochastic neighbor embedding.

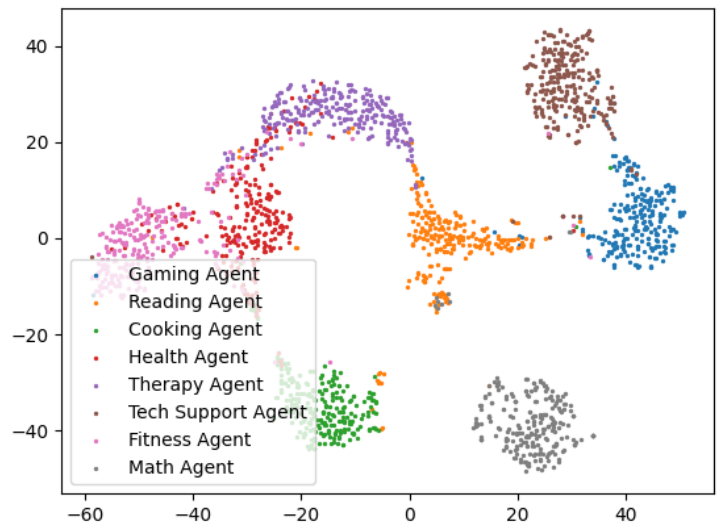

Figure 6: Sentence embeddings from the test split ($N = 2000$) generated by the AgentRec architecture reduced from 768 dimensions to 2 dimensions through t-distributed shochastic neighbor embedding.

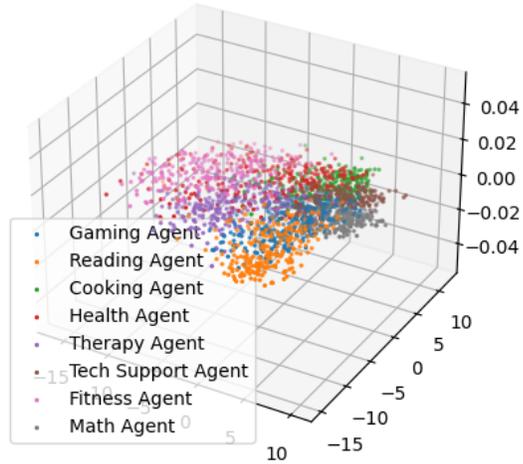

Figure 7: Sentence embeddings from the test split ($N = 2000$) generated by `all-mpnet-base-v2` reduced from 768 dimensions to 3 dimensions through principal component analysis (PCA).

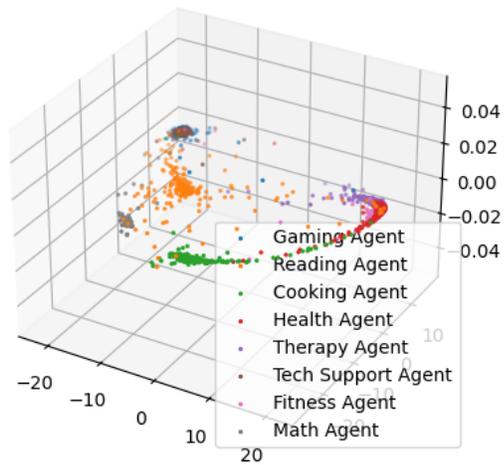

Figure 8: Sentence embeddings from the test split ($N = 2000$) generated by the AgentRec architecture reduced from 768 dimensions to 3 dimensions through principal component analysis (PCA).